\documentclass{article}
\usepackage{amsmath}
\usepackage{amsthm}
\usepackage{graphicx}
\usepackage{tabularx}
\usepackage{makecell}
\usepackage{multirow}
\usepackage{enumitem}
\usepackage{wrapfig}
\usepackage{hyperref}
\usepackage{changepage}
\usepackage{algorithm}
\usepackage{algpseudocode}
\usepackage{todonotes}
\usepackage{booktabs}
\usepackage{tabularx}
\usepackage{multirow}
\usepackage{makecell}
\usepackage[numbers]{natbib}

\newcommand{\methodname}{HM3}
\newtheorem{definition}{Definition}

\title{\methodname: Heterogeneous Multi-Class Model Merging}

\author{
    Stefan Hackmann\\
    \texttt{\small stefan.hackmann@datarobot.com}
}

\date{\vspace{5mm}\today}

\begin{document}
\maketitle
\begin{abstract}
\noindent Foundation language model deployments often include auxiliary ``guard-rail'' models to filter or classify text, detecting jailbreak attempts, biased or toxic output, or ensuring topic adherence. These additional models increase the complexity and cost of model inference, especially since many are also large language models. To address this issue, we explore training-free model merging techniques to consolidate these models into a single, multi-functional model. We propose Heterogeneous Multi-Class Model Merging (\methodname) as a simple technique for merging multi-class classifiers with heterogeneous label spaces. Unlike parameter-efficient fine-tuning techniques like LoRA \citep{hu2021lora}, which require extensive training and add complexity during inference \citep{han2024parameter}\citep{hu2021lora}, recent advancements allow models to be merged in a training-free manner \citep{yadav2023tiesmergingresolvinginterferencemerging}\citep{yu2024languagemodelssupermario}\citep{goddard2024arceesmergekittoolkitmerging}. We report promising results for merging BERT-based guard models \citep{devlin2019bertpretrainingdeepbidirectional}, some of which attain an average F1-score higher than the source models while reducing the inference time by up to 44\%. We introduce \emph{self-merging} to assess the impact of reduced task-vector density, finding that the more poorly performing hate speech classifier benefits from self-merging while higher-performing classifiers do not, which raises questions about using task vector reduction for model tuning.\\

\end{abstract}

\section{Introduction}\label{sec:introduction}
\begin{figure}[!ht]
\centering
\includegraphics[width=1.0\linewidth]{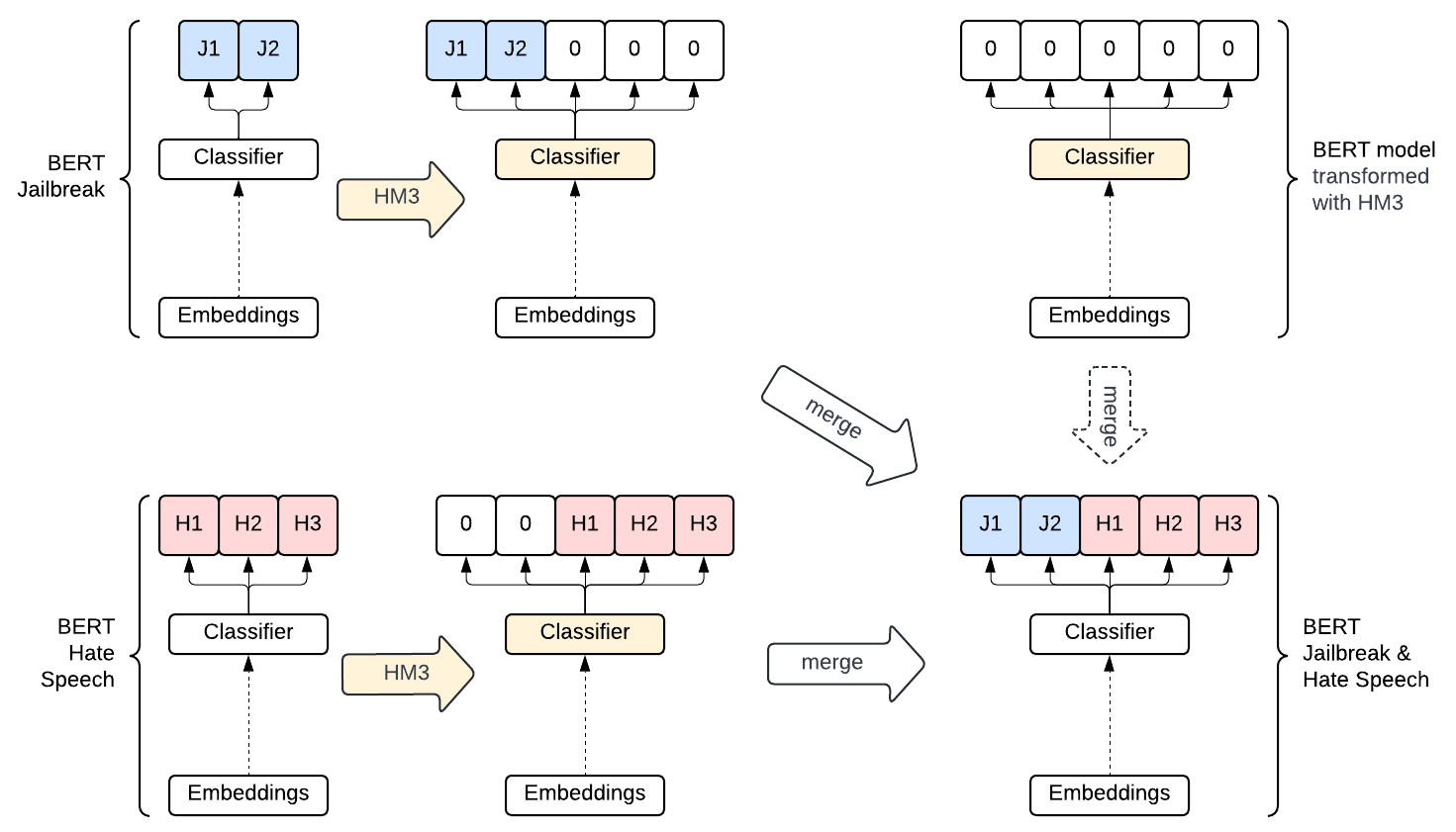}
\caption{Our proposed method \methodname\ transforms a constellation of text classifiers with heterogeneous output dimensions so that they have the same output classifier structure and can be merged. The output layer of the base model is replaced with a classifier that consists only of zeros. Class probabilities of the merged model should be computed group-wise: softmax is applied separately on the two groups (J1, J2) and (H1, H2, H3) so that for each group the probabilities sum up to one. See Section~\ref{sec:our_method} for details.}
\label{fig:guard_merging_simple}
\end{figure}

\begin{table}[!ht]
\centering
\begin{tabularx}{\textwidth}{Xrrrr}
    \toprule
    \textbf{Merged Classifiers}
    & \makecell[l]{\textbf{Original}}
    & \makecell[l]{\textbf{Soup}}
    & \makecell[l]{\textbf{TIES}}
    & \makecell[l]{\textbf{DARE-TIES} \\ {\small (500 Trials)}}\\
    \midrule
    Jailbreak and hate speech
    & 0.768
    & 0.798
    & \underline{0.819}
    & \textbf{0.829}\\
    \addlinespace
    Phishing and sentiment
    & \textbf{0.939}
    & 0.832
    & 0.794
    & \underline{0.925}\\
    \bottomrule
\end{tabularx}
\caption{
The table shows average F1-scores for two case studies - the merger of \texttt{bert-base-uncased}-based models \texttt{jackhhao/jailbreak-classifier} and \texttt{Hate-speech\-CNERG/bert-base-uncased-hatexplain}, and the merger of \texttt{bert-large-uncased}-based models \texttt{ealvaradob/bert-finetuned-phishing} and \texttt{assemblyai/bert-large-uncased-sst2}. We compare the average F1-score of the original models with several merged models. We combine \methodname\ with three merging strategies: Model Soup \citep{wortsman2022model}, TIES \citep{yadav2023tiesmerging}, and model search using repeated merging with DARE-TIES \citep{yu2024language} using different task vector densities. See Subsection~\ref{sub:model_merging} for model merging and Section~\ref{sec:our_method} for our approach to merging heterogeneous text classifiers (\methodname). In the first case study, we see that all merged models outperform the original models on average, with the model resulting from search being the best model. In the second case study, only the model search result has an F1-score comparable to the original models. In both cases, the best models that we found via model search are interesting alternatives to the original models because their performance is at par with the original models or better while we need to run only a single model.}
\label{table:overview}
\end{table}

Production machine learning systems are increasing in complexity. For example, a state-of-the-art generative AI pipeline typically contains multiple sophisticated models working together. A chatbot may have a foundation model like GPT-4 as the main component but be accompanied by a constellation of auxiliary ``guardrail'' models for detecting jailbreak attempts, and biased or toxic output. Other auxiliary models may include application-specific models like sentiment detectors, task-completion detectors, etc. Similarly, mixed-reality action recognition pipelines often have dedicated models for face recognition, keypoint tracking, head tracking, and hand recognition, all of which must run under real-time constraints. Orchestrating a sophisticated constellation of models increases inference complexity and expense. It may also render certain real-time applications infeasible due to exceeding acceptable latency thresholds.

This motivates our investigation into merging multiple models into a single multi-purpose model without losing accuracy on each task compared to the individual models. We consider multi-class classifiers with their own label spaces, e.g., a jailbreak detector might classify text as ``jailbreak’’ or ``no jailbreak’’, while a hate-speech detector classifies text as ``hate-speech’’, ``normal`` or ``offensive’’. These heterogeneous output label spaces pose a unique challenge compared to merging models with homogeneous label spaces, such as a pair of sentiment models, potentially trained on different datasets, each having a ``negative'' and a ``positive'' output label.

While one could approach the task of combining multiple heterogeneous output models using many different techniques like parameter-efficient fine-tuning methods \citep{han2024parameterefficientfinetuninglargemodels} or multi-task learning \citep{yu2024unleashingpowermultitasklearning}, in this work, we investigate training-free model merging methods, see Subsection~\ref{sub:model_merging}. Such methods are especially attractive because they do not require further training, which can be prohibitively expensive. Model merging can run on CPU in a matter of minutes. Recent advances in training-free model merging methods have shown promising results in merging pre-trained models \citep{yadav2023tiesmerging}\citep{yu2024language}. We ask the question \emph{``Can homogeneous model merging techniques be adapted to merging text classifiers with heterogeneous label spaces without notable loss of accuracy?''} We conclude in the positive (see Table~\ref{table:overview}). We name our method Heterogeneous Multi-Class Model Merging, abbreviated as \methodname.

In Section~\ref{sec:background}, we review model merging techniques especially relevant to the techniques used in this work. Section~\ref{sec:our_method} gives a description of the adaptations we contribute towards merging models with heterogeneous label spaces and Model Search. Section~\ref{sec:case_studies} reviews our methods and results merging guard models for LLM deployments.

\section{Background}\label{sec:background}
\subsection{Model Merging}\label{sub:model_merging}
Model merging is a class of techniques to merge model capabilities from different models without any further training.
For a broad overview of these techniques, we refer to \citep{li2023deep}.
Model Soup \citep{wortsman2022model} is a straightforward but powerful strategy, which involves taking the weighted average of the weights of several fine-tuned models.
It is an important baseline method because of its simplicity.

\citet{ilharco2023editing} define ``task vectors'' to be the weights of a base model subtracted from those of a fine-tuned version of the same model.
Task vectors encode the task-specific capabilities of the fine-tuned model and are an important concept for TIES-merging \citep{yadav2023tiesmerging}.
TIES (TrIm and Elect Sign) deals with the problem of parameter interference and consists of three steps:

\begin{enumerate}
    \item Keep the top-$k$\% largest task vector values of each to-be-merged model and set all other values to zero.
    \item For each task vector parameter $p$, the merged model's $m$ assigned sign $\gamma_m^p$ is the sign of the sum of the task vector values $\tau_t^p$ across the task vectors: $\gamma_m^p = \operatorname{sgn}(\sum_{t=1}^N \tau_t^p)$, where $N$ is the number of models to be merged.
    \item Merge the task vectors parameter-wise by computing the average of all non-zero values that have the same sign as $\gamma_m^p$.
\end{enumerate}

\citet{yu2024language} introduce the pre-processing technique DARE, short for Drop And REscale, where a fraction of randomly chosen values from task vectors are dropped and the remaining values linearly rescaled.
DARE is found to preserve, and in certain cases increase, task performance when combined with a merging strategy like Model Soup or TIES.
Note that we will mostly talk about task vector densities ($1-\mbox{drop rate}$) instead of drop rates.

\section{Related Work}\label{sec:related_works}
\subsection{Ensembling}\label{sub:ensembling}
Model merging is distinct from model ensembling \citep{Breiman1996}\citep{9893798} in that it produces a single model rather than combining the outputs of several models.
Ensembling text classifiers has been studied by various authors, for example \citet*{MOHAMMED20228825} who propose an ensemble deep learning framework for text classification and compare the proposed ensemble with other methods.
However, these methods do not support heterogeneous text classification tasks with distinct classification semantics or output label spaces.
LLMs like GPT-4 \citep{openai2024gpt4}, Claude 3 \citep{anthropic2024claude3} and many others can be prompted to act as classifiers and can be ensembled, but they are slow and expensive to run.

\subsection{Multi-Task Learning}\label{sub:multi-task-learning}
Model merging is related to the wider and established field of multi-task learning (MTL) \citep{CMU-CS-97-203}, where machine learning models are trained to complete multiple, related tasks in order to improve the model's performance across one or more of those tasks.
Model merging has a comparable effect but is conducted after training.
\citet{zhang2021surveymultitasklearning} provide a survey on MTL.
Not surprisingly, MTL can also be regarded as multi-objective optimization \citep{sener2019multitasklearningmultiobjectiveoptimization}. 
\citet{bhattacharjee2022multendtoendmultitasklearning} build a transformer architecture tailored to multi-task learning, named MulT.
Recently, \citet{yang2024adamergingadaptivemodelmerging} introduced Adaptive Model Merging (AdaMerging), which learns model merging coefficients in an unsupervised manner from unlabeled multi-task test data.

\subsection{LoRA}\label{sub:lora}
An important alternative technique for extending model capabilities without full fine-tuning is low-rank adaption of LLMs (LoRA) \citep{hu2021lora}.
Instead of fine-tuning the weights of the base model, LoRA injects a reduced number of trainable weights into each layer of the extended LLM while keeping the original models' weights fixed.
LoRA increases the number of parameters required for inference, but it has a negligible impact on latency because the additional weights can be evaluated in parallel.
Two popular systems for serving LoRA-adapted LLMs are vLLM, introduced by \citet{kwon2023efficient}, and S-LoRA, introduced by \citet{sheng2024slora}.
S-LoRA can handle a large collection of concurrent LoRA adapters, significantly increasing throughput compared to vLLM, but does not enable multi-task inference.
In this work, we focus on pushing the boundaries of training-free model merging methods, which do not require additional parameters and enable multi-task inference.

\section{Heterogeneous Multi-Class Model Merging}\label{sec:our_method}
In our case-studies, we use Hugging Face as a resource for text classifiers.
We merge models that are fine-tuned versions of the same base model but adapted to heterogeneous label spaces.
The merging strategies applied here require that the merged models have the same architecture.
Some merging strategies like TIES also require the common base model from which the fine-tuned models were fine-tuned in order to compute task vectors.
The base models, fine-tuned models, and merged models can all have different output spaces and semantics.
In order to be able to merge these models, we expand the final classifier layer of the models that we want to merge with zeros and adjust the base model accordingly, see Algorithm~\ref{alg:method}.
Formally, we can describe the pre-processing step HM3 as follows.

\begin{algorithm}
\caption{Transform $N$ models and their base model with \methodname}\label{alg:method}
\begin{algorithmic}[1]
\Statex \textbf{Input}: heterogeneous models $(m_1, \ldots, m_N)$ to be merged and corresponding base model $b$
\For{model $i$ to be merged}
    \State Set $K_i \gets$ number of outputs of the model
\EndFor
\For{model $i$ to be merged}
    \State Set $K_\text{before} \gets \sum_{j=1}^{i-1} K_j$
    \State Set $K_\text{after} \gets \sum_{j=i+1}^N K_j$
    \State Set $\tilde{m}_i \gets m_i$ with an output layer that has $K_\text{before}$ additional zeros to the left and $K_\text{after}$ additional zeros to the right 
\EndFor
\State Set $\tilde{K} \gets \sum_i K_i$
\State Set $\tilde{b} \gets b$ with a zero tensor as the new output layer that returns $\tilde{K}$ zeros and has the same dimensions as the output layers of the transformed models
\Statex \textbf{Output:} homogeneous set of models $(\tilde{m}_1, \ldots, \tilde{m}_N)$ and new base model $\tilde{b}$
\end{algorithmic}
\end{algorithm}

\begin{definition}\label{def:hm3}
Let $\mathcal{F} = \{f_1, \ldots, f_N\}$ be a collection of $N$ 
text classifiers, where $f_i$ maps a text $x$ to $K_i$ classes:
\begin{equation}
f_i: x \mapsto \left(p_{i,1}, \ldots, p_{i,K_i}\right),
\end{equation}
such that $\sum_{j=1}^{K_i} p_{i,j} = 1$, for all $i \in \{1, \ldots, N\}$. Here, $p_{i,j}$ is the probability that the input belongs to the $j$-th class of the $i$-th classifier. Each $f_i$ consists of a tokenization step $\tau$ that maps the text $x$ to tokens $(x_1, \ldots, x_T)$, a forward pass by the model $m_i$ that converts the tokens to logits
\begin{equation}
m_i: (x_1, \ldots, x_T) \mapsto (l_{i,1}, \ldots, l_{i,K_i})
\end{equation}
and a step that transforms the logits to class probabilities using the $\operatorname{softmax}$ function. We write
\begin{equation}
f_i = \operatorname{softmax} \circ\, m_i \circ \tau.
\end{equation}
The corresponding HM3-expanded classifier $\tilde{f}_i$ has additional zeros for the outputs of all other classifiers from $\mathcal{F}$, therefore $\tilde{K} = \sum_{i=1}^N K_i$ output labels, and is given by
\begin{equation}\label{eqn:classifier}
\tilde{f}_i := \operatorname{softmax}^* \circ\, \tilde{m}_i \circ \tau,
\end{equation}
\begin{equation}
\tilde{m}_i: (x_1, \ldots, x_T) \mapsto \left(0_{K_1}, \ldots, 0_{K_{i-1}}, l_{i,1}, \ldots, l_{i,K_i}, 0_{K_{i+1}}, \ldots, 0_{K_N}\right),
\end{equation}
where $(x_1, \ldots, x_T)$ are $T$ tokens resulting from the tokenization step, $0_n$ denotes a segment consisting of $n$ zeros and $\operatorname{softmax}^*$ is the $\operatorname{softmax}$ function applied to each segment individually:
when $l_i = l_{i,1}, \ldots, l_{i,K_i}$ is the segment corresponding to the output of model $m_i$, then
\begin{equation}
\operatorname{softmax}^*: (l_1, \ldots, l_N) \mapsto \left(\operatorname{softmax}(l_1), \ldots \operatorname{softmax}(l_N)\right).
\end{equation}
To generate the HM3-transformed base model $\tilde{m}_b$, we replace the output layer of the base model $m_b$ with a zero-tensor that generates $\tilde{K}$ outputs.
\end{definition}

To merge a collection of models $\{m_1, \ldots, m_N\}$ with heterogeneous label spaces but common base model $m_b$, we can apply HM3 as defined in Definition~\ref{def:hm3} to generate the transformed models $\{\tilde{m}_1, \ldots, \tilde{m}_N\}$ and $\tilde{m}_b$. Those can be merged with standard techniques like Model Soup or TIES into a new model $\tilde{m}$. The resulting merged text classifier $\tilde{f}$ is then given by
\begin{equation}
\tilde{f} = \operatorname{softmax}^* \circ\, \tilde{m} \circ \tau,
\end{equation}
where $\operatorname{softmax}^*$ and $\tau$ are defined as in Definition~\ref{def:hm3}.
See Algorithm~\ref{alg:method} for a compact representation of HM3 and Algorithm~\ref{alg:merge} for its application to model merging.

\begin{algorithm}
\caption{Model Merge with \methodname}\label{alg:merge}
\begin{algorithmic}[1]
\Statex \textbf{Input}: heterogeneous models $m_i$ to be merged and corresponding base model, $i = 1, \ldots, N$
\State Transform models with \methodname
\State Prepare a merge with Model Soup, TIES or DARE-TIES
\State Merge models to form $\tilde{m}$ with mergekit
\State Evaluate $\tilde{m}$ on 3000 test samples per test dataset
\Statex \textbf{Output:} Merged model and evaluation data
\end{algorithmic}
\end{algorithm}

Note that the configuration details for output labels of open-source text classifier are sometimes missing or not meaningful, for instance, labels may be called ``label0'', ``label1'', etc.
When expanding the output classifiers, we replace such uninformative output labels and ensure that there are no duplicate labels: e.g., when ``normal'' is used by several models.

\begin{algorithm}
\caption{Model Search with \methodname}\label{alg:search}
\begin{algorithmic}[1]
\Statex \textbf{Input}: heterogeneous models $m_i$ to be merged and corresponding base model, $i = 1, \ldots, N$
\State Initialize best model as $\tilde{m}_{\text{best}}$
\State Transform models with \methodname
\For{$i = 1$ to $500$}
    \State Sample a task vector density from $\text{Beta}(1.2, 2)$
    \State Prepare a merge with DARE-TIES
    \State Merge models to form $\tilde{m}$ with mergekit
    \State Evaluate $\tilde{m}$ on 600 validation samples per test dataset
    \If{$\tilde{m}$ is better than $\tilde{m}_{\text{best}}$}
        \State Evaluate $\tilde{m}$ on 1000 test samples per test dataset
        \State Update $\tilde{m}_{\text{best}} \gets \tilde{m}$
    \EndIf
\EndFor
\Statex \textbf{Output:} Best merged model $\tilde{m}_{\text{best}}$ and evaluation data
\end{algorithmic}
\end{algorithm}

\section{Case Studies}\label{sec:case_studies}
\begin{wrapfigure}{r}{0.5\textwidth}
  \centering
  \includegraphics[width=0.48\textwidth]{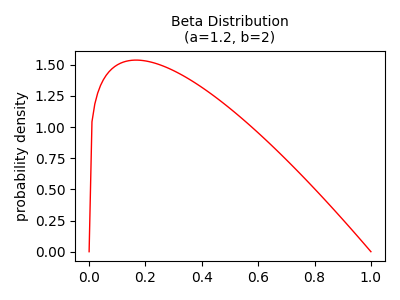}
  \caption{For model search, we sample task vector densities that are used by DARE-TIES after pre-processing with HM3. We use a Beta distribution for sampling that is skewed to the right to quickly explore the interesting cases with low but non-zero density. We select a distribution with zero density at zero as dropping all task vector values is equivalent to undoing all of the fine-tuning. Similarly, there is no variability in outcomes when the density is 1 and the merge algorithm is deterministic.}
  \label{fig:beta_distribution}
\end{wrapfigure}

\begin{figure}[!ht]
\centering
\includegraphics[width=1.0\linewidth]{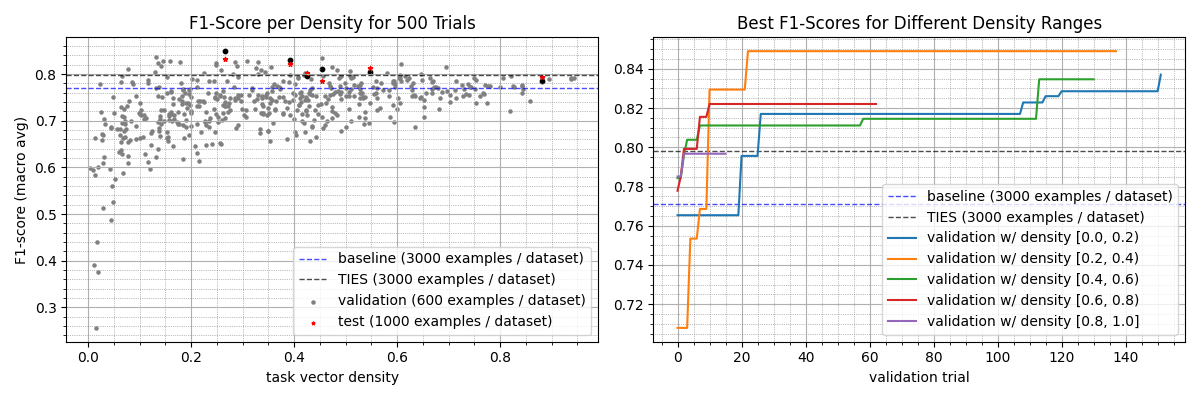}
\caption{Both plots show F1-scores resulting from 500 guard models. We merged a jailbreak classifier with 2 output labels, \texttt{jackhhao/jailbreak\-classifier}, and a hate speech classifier with 3 output labels, \texttt{Hate-speech\-CNERG/bert-base-uncased-hatexplain}, into a single classifier with 5 output labels using \methodname\ followed by DARE-TIES, where we use the same but changing task-vector densities for both input models. The horizontal blue lines show the average macro F1-score of the original models. We evaluated 3000 test examples per dataset to compute the baseline. The stars mark additional test results for new best merged models found during model search, see Algorithm~\ref{alg:search}, and the bold dots are the corresponding validation results that triggered the additional test. For densities close to 1, merging generally improves the overall performance a bit. For densities close to 0, the merged models frequently fail to classify examples correctly. Surprisingly, the best models are generated with quite low task-vector densities.}
\label{fig:bert-base-uncased__search}
\end{figure}

We conducted two case studies where we merge fine-tuned text classifiers with mergekit \citep{goddard2024arcee} that have been modified with \methodname, see Section~\ref{sec:our_method}. We explore several merging strategies: Model Soup, TIES, and DARE-TIES, see Subsection~\ref{sub:model_merging}. When using \methodname\ with DARE-TIES, we sample the task vector density from a beta distribution with $\alpha=1.2$ and $\beta=2$, see Figure~\ref{fig:beta_distribution}.
The scatter plot in Figure~\ref{fig:bert-base-uncased__search} is one example that shows how test result scores can depend on the task vector density. The elevated variance in the distribution of scores justifies the implementation of a search process, see Algorithm~\ref{alg:search}.

In our case studies, we explore text classifiers that are used as guard models because guard models are of paramount importance in the context of LLM moderation.

\subsection{LLM Moderation}
\label{sub:moderation}
Data engineers, machine learning engineers and prompt engineers go to great length to make LLMs safer and yet, we are still far from done. See, for instance, \citet{naveed2024comprehensive} for an introduction to the field of LLMs in general and \citet{cui2024risk} for related risks. A starting point for risk mitigation is the selection of training data. Anthropic puts an emphasis on safety and quality of training data \citep{wei2023jailbroken}, for instance. Similarly, you could try to prevent harmful data from entering your RAG-system. 
To make models less vulnerable to certain types of attacks and make them respond more like humans would prefer, fine-tuning \citep{naveed2024comprehensive} and reinforcement learning from human feedback (RLHF) \citep{christiano2023deep} are used to further improve pre-trained models and make them safer. Additional instructions in the model's context can also provide an extra bit of safety \citep{sahoo2024systematic}\citep{kamruzzaman2024prompting}, for instance, you can reduce the overall risk by instructing a model to stay on topic because many attacks will not be aligned with the topic. All those efforts are important but still do not prevent LLMs from producing harmful content in certain situations. Here, we are particularly interested in another layer of protection that we aim to apply to make models safe enough to use them in a professional or educational environment: specialized guard models are used to analyze the data that is entering a model or is produced by a model \citep{dong2024building}. Those guard models are often text classifiers but could also be image classifiers or LLMs themselves. Typical examples are jailbreak, toxic speech and phishing classifiers.\footnote{You can find plenty of useful classifiers on Hugging Face, for instance:\\
\texttt{https://huggingface.co/\texttt{Hate-speech-CNERG/bert-base-uncased-hatexplain}}\\
\texttt{https://huggingface.co/\texttt{jackhhao/jailbreak-classifier}}\\
\texttt{https://huggingface.co/ealvaradob/bert-finetuned-phishing}
}

\subsection{Case Study 1}\label{sub:case-study-1}
\begin{table}[!ht]
\centering
\begin{tabularx}{\textwidth}{lXl}
    \toprule
    \textbf{Category} & \textbf{Model} & \textbf{Labels} \\
    \midrule
    Jailbreaks & \texttt{jackhhao/jailbreak-classifier} & 
    \makecell[l]{0) benign \\ 1) jailbreak} \\
    \addlinespace
    Toxic content & \makecell[l]{\texttt{Hate-speech-CNERG}/\\ \texttt{bert-base-uncased-hatexplain}} & 
    \makecell[l]{0) hate speech \\ 1) normal \\ 2) offensive} \\
    \bottomrule
\end{tabularx}
\caption{Classifiers fine-tuned on BERT base uncased that we use in our first merging case study. While the jailbreak classifier has two output labels, the hate speech classifier has three.}
\label{table:fine-tuned-versions-of-bert-base-uncased}
\end{table}

In our first case study, we merge two models that are fine-tuned versions of BERT base uncased, see Table~\ref{table:fine-tuned-versions-of-bert-base-uncased}. Note that those models have different numbers of output labels. While \texttt{jackhhao/jailbreak-classifier} has two output labels, \texttt{Hate-speech-CNERG/bert-base-uncased-hatexplain} has three. The resulting model, when merging with \methodname, therefore has five output labels, see Section~\ref{sec:our_method} for details.

\begin{table}[!ht]
\centering
\begin{tabularx}{\textwidth}{Xll}
    \toprule
    \textbf{Model} & \textbf{Test} & \textbf{Expected Label} \\
    \midrule
    \multirow{4}{=}{\texttt{jackhhao/jailbreak-classifier}} 
        & MMLU & benign \\
        & jailbreaks & jailbreak \\
        & hate speech & benign \\
        & offensive & benign \\
    \midrule
    \multirow{3}{=}{\makecell[l]{\texttt{Hate-speech-CNERG/}\\\texttt{bert-base-uncased-hatexplain}}}
        & MMLU & normal \\
        & hate speech & hate speech \\
        & offensive & offensive \\
    \bottomrule
\end{tabularx}
\caption{Our test datasets and the corresponding expected labels. We have only one expected class per test dataset and use the same number of examples per dataset if not mentioned otherwise. We created the MMLU test using the \texttt{cais/mmlu} dataset from Hugging Face. The jailbreaks test is based on the data from the GitHub project \url{https://github.com/verazuo/jailbreak_llms}. The hate speech and offensive tests are based on the \texttt{hatexplain} dataset from Hugging Face: we allocate a text to the hate speech or offensive test dataset depending on whether there are more instances of hate speech or offensive speech in the given text.}
\label{table:bert-base-uncased-expected-labels-for-tests}
\end{table}

We decided to use a test setup where we have at least one dedicated dataset for each label, where this particular label should have the highest probability. In other words, every example from this dataset is a positive example for the corresponding label, for instance, we have a jailbreak dataset, where each entry is a combination of a jailbreak with a question that should not be answered by an LLM. The guard model is expected to mark every example as a jailbreak. A dataset that we are frequently using to test the negative case, for instance, not a jailbreak or not a phishing attempt, is MMLU because of its great diversity of texts.\footnote{\texttt{https://huggingface.co/datasets/cais/mmlu}} In some cases, we use the datasets that have been used for fine-tuning. Note that we don't aim to provide an independent evaluation of the fine-tuned models but want to quantify the performance impact of merging.
\begin{figure}[!ht]
\centering
\includegraphics[width=1.0\linewidth]{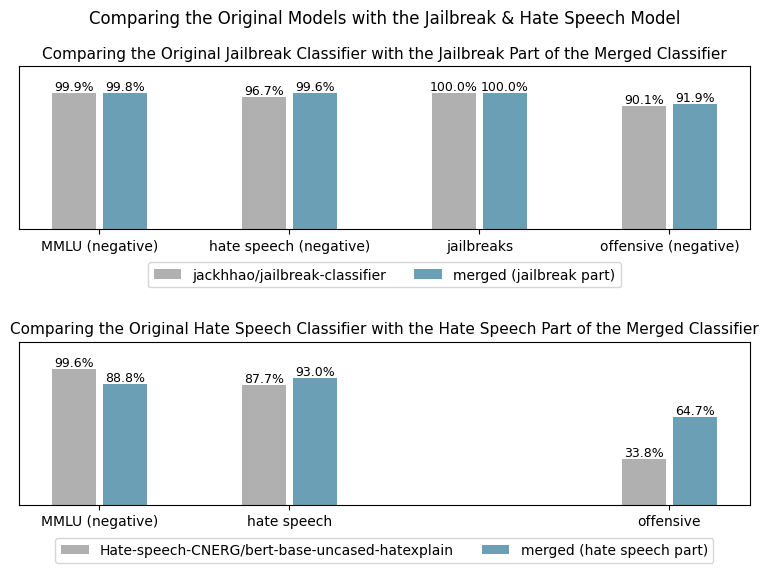}
\caption{Here we compute the F1-score of the original classifier and the merged model resulting from model search for each dataset individually. The merged model is the model with the best validation score from model search. Note that we ``cross-check'' the merged model where possible, for instance, we test the jailbreak part of the merged model using the dataset with positive hate speech examples because we like to see that the merged model labels those as no jailbreaks. \emph{Positive jailbreak examples often contain problematic language and therefore we omit testing the hate speech part with positive jailbreak examples.} See Table~\ref{table:bert-base-uncased-expected-labels-for-tests} for a list of expected labels, Appendix~\ref{sub:accuracies} for a complete collection of model accuracy comparisons, and Appendix~\ref{sub:testing} for more insights into our testing process.}
\label{fig:bert-base-uncased-jailbreak-hatespeech__dare-ties__f1-score}
\end{figure}

A model search using \methodname\ followed by DARE-TIES gives the best merged model, see Algorithm~\ref{alg:search} for model search. In the first case study, where we merged a jailbreak and a hate speech classifier, the merged models even outperform the original models, see Table~\ref{table:overview}.

\begin{table}[!ht]
\centering
\begin{tabularx}{\textwidth}{Xrrr}
    \toprule
    & Original & Merged & Reduction \\
    \midrule
    Load duration/[min] & 68 & 35 & 48\% \\
    Inference duration/[min] & 40 & 25 & 37\% \\
    \midrule
    Total/[min] & 108 & 60 & 44\% \\
    \bottomrule
\end{tabularx}
\caption{Combined runtime of two individual models compared to the runtime of one merged model. This is based on data collected from a model search, combining \texttt{jackhhao/jailbreak-classifier} and \texttt{Hate-speech\-CNERG/bert-base-uncased-hatexplain} into a new text classifier.}
\label{tab:model-runtime-comparison}
\end{table}

\subsection{Case Study 2}\label{sub:case-study-2}
\begin{table}[!ht]
\centering
\begin{tabularx}{\textwidth}{lXl}
    \toprule
    \textbf{Category} & \textbf{Model} & \textbf{Labels} \\
    \midrule
    Phishing & \texttt{ealvaradob/bert-finetuned-phishing} & 
    \makecell[l]{0) non-phishing \\ 1) phishing} \\
    \addlinespace
    Sentiment & \texttt{assemblyai/bert-large-uncased-sst2} & 
    \makecell[l]{0) negative \\ 1) positive} \\
    \bottomrule
\end{tabularx}
\caption{In our second case study, we merge the phishing detector \texttt{ealvaradob/bert-finetuned-phishing} and the sentiment classifier 
\texttt{assemblyai/bert-large-uncased-sst2}.}
\label{table:test-datasets-for-bert-base-uncased-models}
\end{table}

The underlying base model for our first study was BERT base uncased, a deep bidirectional transformer with 110M parameters. In our second case study, we merge two models that are fine-tuned versions of BERT large uncased, the larger sibling with 340M parameters \citep{devlin2019bertpretrainingdeepbidirectional}.

\begin{table}[!ht]
\centering
\begin{tabularx}{\textwidth}{Xll}
    \toprule
    \textbf{Model} & \textbf{Test} & \textbf{Expected Label} \\
    \midrule
    \multirow{2}{*}[-0.5ex]{\makecell[l]{\texttt{ealvaradob/bert-finetuned-phishing}}} 
        & non-phishing & non-phishing \\
        & phishing & phishing \\
    \midrule
    \multirow{2}{*}[-0.5ex]{\makecell[l]{\texttt{assemblyai/bert-large-uncased-sst2}}}
        & negative & negative \\
        & positive & positive \\
    \bottomrule
\end{tabularx}
\caption{When testing \texttt{ealvaradob/bert-finetuned-phishing}, \texttt{assemblyai/ bert-large-uncased-sst2} and the corresponding merged models, we have one dedicated dataset for every label. Where possible, we use random samples from the datasets that have also been used for fine-tuning because we want to use examples where the fine-tuned models are supposed to work well and then evaluate whether the merged models perform comparatively. We generated the ``non-phishing'' and the ``phishing'' test from the \texttt{ealvaradob/phishing-dataset} dataset from Hugging Face and the ``negative'' and ``positive'' test from \texttt{stanfordnlp/sst2} dataset from Hugging Face.}
\label{tab:bert-large-uncased-expected-labels-for-tests}
\end{table}

Overall, in the second case study, the merged models perform worse than the original models but model search using \methodname\ followed by DARE-TIES gives comparable results. Appendix~\ref{sub:confusion_matrices} shows confusion matrices for both case studies.

\begin{figure}[!ht]
\centering
\includegraphics[width=1.0\linewidth]{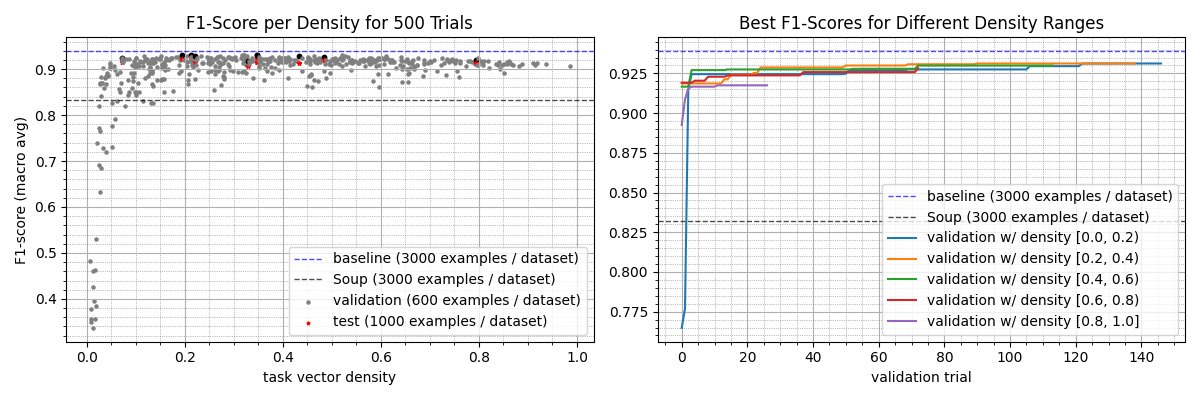}
\caption{Merge of the phishing detector \texttt{ealvaradob/bert-finetuned\-phishing} with the sentiment classifier \texttt{assemblyai/bert-large-uncased-sst2} using DARE-TIES. Like in Figure~\ref{fig:bert-base-uncased__search}, we find the best as well as the worst F1-scores for low densities. The best merged model has roughly the F1-score as the individual models on average, see Table~\ref{table:overview}}
\label{fig:selfbert-large-uncased-phishing-sentiment__searchmerge_jailbreak}
\end{figure}
\begin{figure}[!ht]
\centering
\includegraphics[width=1.0\linewidth]{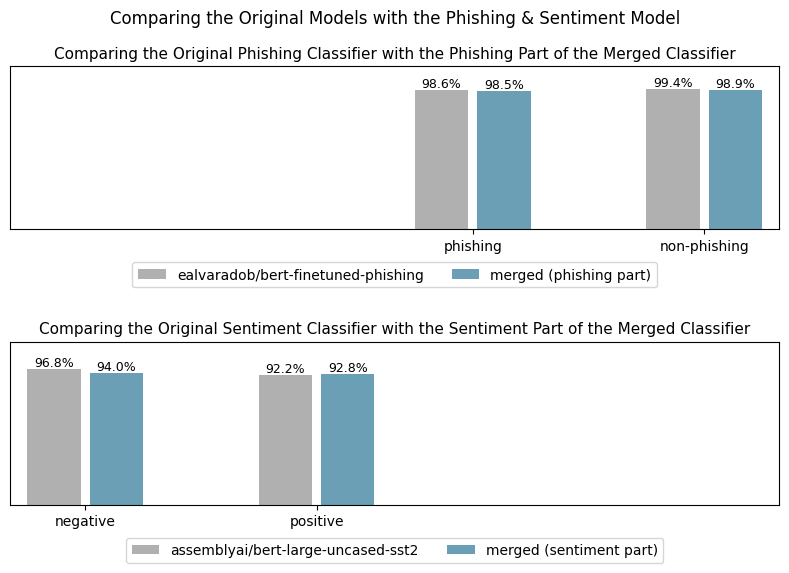}
\caption{When merging the phishing and the sentiment classifier using model search and \methodname\ followed by DARE-TIES, the merged model shows only a negligible performance degradation. Here, we did not test the ``phishing part'' with sentiment examples nor did we test the ``sentiment part'' of the merged model with phishing examples. Refer to Table~\ref{tab:bert-large-uncased-expected-labels-for-tests} for a list of expected labels, Appendix~\ref{sub:accuracies} for a complete collection of model accuracy comparisons, and Appendix~\ref{sub:testing} for more insights into our testing process.}
\label{fig:bert-large-uncased-phishing-sentiment__dare-ties__f1-score}
\end{figure}

\subsection{Self-Merging}\label{sub:selfmerging}
The best \methodname-based models have been produced with DARE-TIES, using a low density. We were asking ourselves \emph{``Can random resetting of task-vectors without merging with a different model already improve a fine-tuned model or is the low density merely beneficial because this reduces the number of conflicts while merging and improved performance stems from merging with another model?''}
\begin{table}[!ht]
\centering
\begin{tabularx}{\textwidth}{lXrr}
    \toprule
    \textbf{Category}
    & \textbf{Model} 
    & \textbf{Original}
    & \textbf{Self-Merge}\\
    \midrule
    Jailbreak
    & \texttt{jackhhao/jailbreak-classifier}
    & \textbf{0.924}
    & 0.844 \\
    \addlinespace
    Hate speech
    & {\makecell[l]{\texttt{Hate-speech-CNERG/}\\\texttt{bert-base-uncased-hatexplain}}}
    & 0.619
    & \textbf{0.729} \\
    \addlinespace
    Phishing
    & {\makecell[l]{\texttt{ealvaradob/}\\
    \texttt{bert-finetuned-phishing}}}
    & \textbf{0.981}
    & 0.944 \\
    \addlinespace
    Sentiment
    & {\makecell[l]{\texttt{assemblyai/}\\
    \texttt{bert-large-uncased-sst2}}}
    & \textbf{0.896}
    & 0.774 \\   
    \bottomrule
\end{tabularx}
\caption{Average F1-scores for text classifiers and corresponding self-merges. The self-merge is the best result from a model search using DARE and 500 iterations.}
\label{table:selfmerge}
\end{table}
We applied a simple trick to answer this question and used HM3 with model search, where we merged a model with itself using DARE-TIES and varying task-vector densities. We see in Table~\ref{table:selfmerge} that the self-merge is performing worse than the original model for jailbreaks but performing better in the hate speech category, see Appendix~\ref{sub:selfmerging_results} for details.
A potential explanation for the outperformance of the self-merged model in the hate speech case could be that the model is overfitting to the training data and therefore benefits from undoing the training by resetting to base model parameters.

\section{Limitations and Future Directions}\label{sec:limitations_and_future_directions}
The evaluated BERT-models can process only 512 tokens at once. Longer sequences should be split and evaluated separately in practice.

It would have been interesting to merge DeBERTa \citep{he2021debertadecodingenhancedbertdisentangled} models and try AdaMerging \citep{yang2024adamergingadaptivemodelmerging}. This is something we may add at a later point.

We discussed two interesting cases in detail, however, not every constellation of models can easily be merged into a well-performing single model.
\begin{figure}[!ht]
    \centering
    \includegraphics[width=1.0\linewidth]{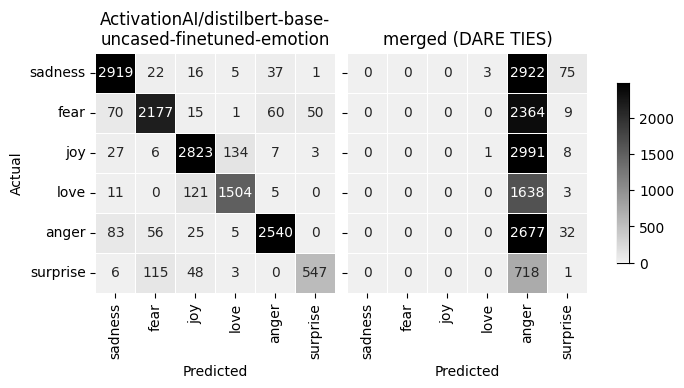}
    \caption{A negative example for model merging where the merged model misclassifies most texts while the original emotion detector is performing quite well.}
    \label{fig:distilbert-base-uncased-emotion-toxic-jailbreak__dare-ties__emotion-detection__confusion_matrices__simple}
\end{figure}
For instance, when we merged three fine-tuned versions of DistilBert base uncased, most texts that test the emotion classifier were classified as ``anger'', see Figure~\ref{fig:distilbert-base-uncased-emotion-toxic-jailbreak__dare-ties__emotion-detection__confusion_matrices__simple}.\footnote{See\\
\texttt{https://huggingface.co/ActivationAI/distilbert-base-uncased-finetuned-emotion}\\
\texttt{https://huggingface.co/martin-ha/toxic-comment-model}\\
\texttt{https://huggingface.co/Necent/distilbert-base-uncased-detected-jailbreak}} We observed this behavior several times and would like to investigate the root cause at a later point.

The field of text classifiers is rather bifurcated with some but not all relevant classifiers available for each model architecture. From the perspective of model merging, it would be favorable to have a larger set of dedicated classifiers for the same architecture that could be merged as needed.

Finally, \methodname\ is not restricted to text classifiers. The same technique should help to merge image classifiers, for instance. It would be very interesting to see how well \methodname\ is performing in other contexts, for instance, in combination with AdaMerging that shows good results when merging image classifiers.

\section{Conclusion}\label{sec:conclusion}
We demonstrated that Heterogeneous Multi-Class Model Merging (\methodname) can be used to merge several text classifiers with different output labels into a single classifier such that the resulting classifier supports all labels. We found that using one such merged model instead of several individual models requires significantly less compute time. This has a positive impact on costs, energy consumption and the environment. We observe that the quality of the merged model is often comparable to the original models, and sometimes even better. We get the best merging results with \methodname\ followed by DARE-TIES and a low task-vector density. Synergies between different merged models that have been fine-tuned on different training data cannot be the only reason for the increased performance we see in some situations because we can produce higher-performing models by simply merging a model with itself, using a low task-vector density.

\section{Acknowledgments}
This work would not have been possible without the support of DataRobot and the help of the OCTO team. I want to especially thank my coworkers Dr. Michael Schmidt, Alexander Conway, Dr. Debadeepta Dey and Mark Steadman for their invaluable support while undertaking this research.

\bibliography{references}
\newpage
\section{Appendix}\label{sec:appendix}

\subsection{Comparing Model Accuracy}\label{sub:accuracies}

\begin{figure}[H]
\centering
\includegraphics[width=1.0\linewidth]{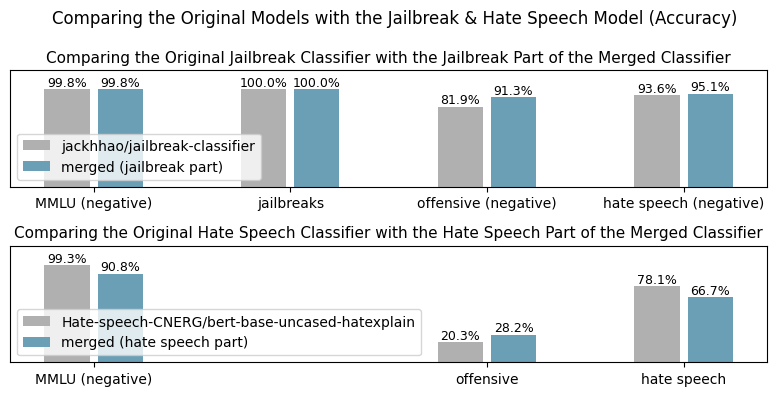}
\caption{Comparing the accuracy of the original classifiers with the merged classifier generated with HM3 and Model Soup.}
\label{fig:bert-base-uncased-jailbreak-hatespeech__soup__accuracy}
\end{figure}

\begin{figure}[H]
\centering
\includegraphics[width=1.0\linewidth]{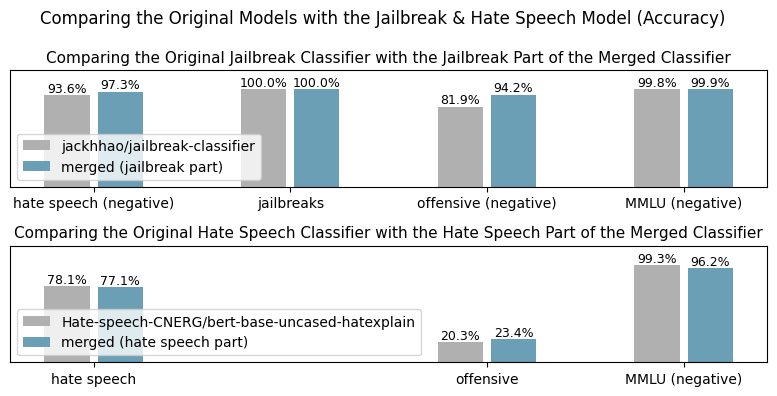}
\caption{Comparing the accuracy of the original classifiers with the merged classifier generated with HM3 and TIES.}
\label{fig:bert-base-uncased-jailbreak-hatespeech__ties__accuracy}
\end{figure}

\begin{figure}[H]
\centering
\includegraphics[width=1.0\linewidth]{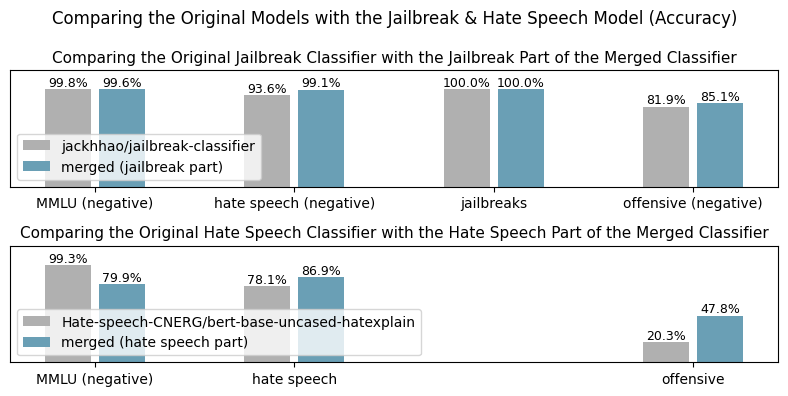}
\caption{Comparing the accuracy of the original classifiers with the merged classifier generated with HM3 and DARE-TIES (model search).}
\label{fig:bert-base-uncased-jailbreak-hatespeech__dare-ties__accuracy}
\end{figure}

\begin{figure}[H]
\centering
\includegraphics[width=1.0\linewidth]{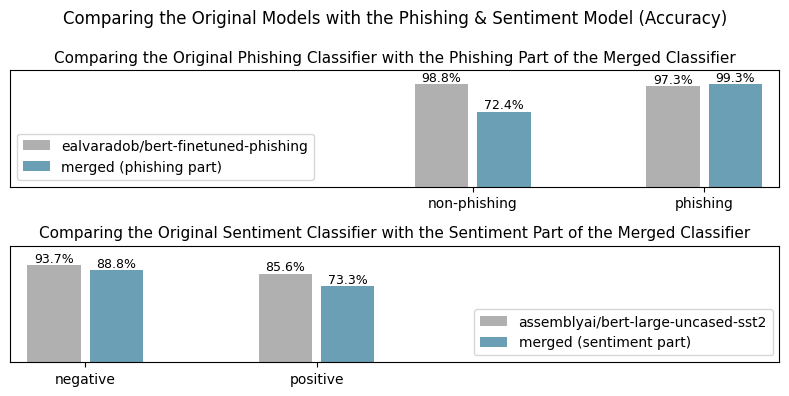}
\caption{Comparing the accuracy of the original classifiers with the merged classifier generated with HM3 and Model Soup.}
\label{fig:bert-large-uncased-phishing-sentiment__soup__accuracy}
\end{figure}

\begin{figure}[H]
\centering
\includegraphics[width=1.0\linewidth]{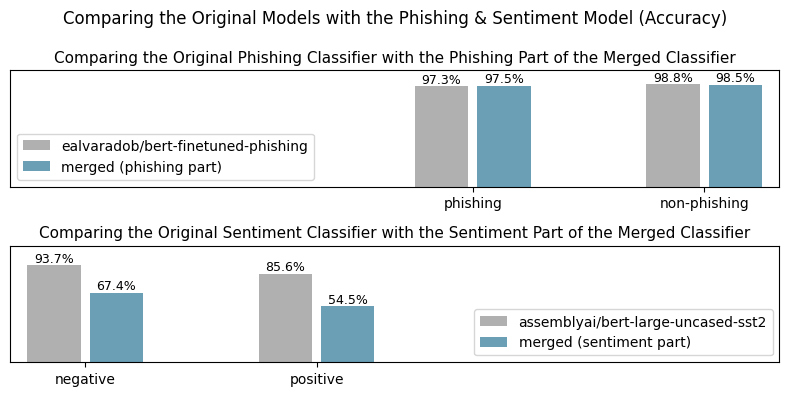}
\caption{Comparing the accuracy of the original classifiers with the merged classifier generated with HM3 and TIES.}
\label{fig:bert-large-uncased-phishing-sentiment__ties__accuracy}
\end{figure}

\begin{figure}[H]
\centering
\includegraphics[width=1.0\linewidth]{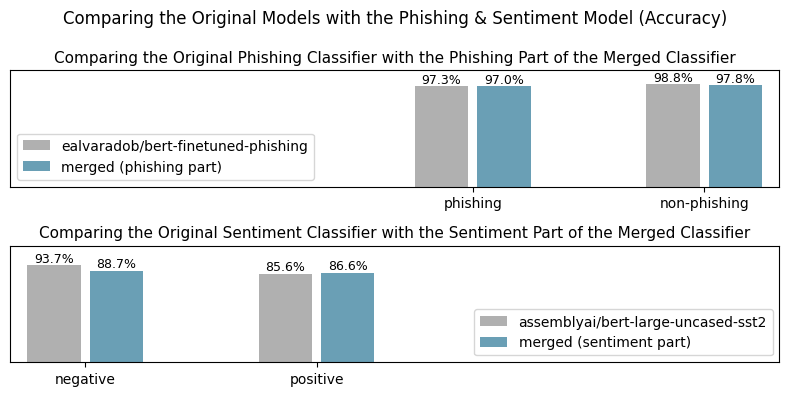}
\caption{Comparing the accuracy of the original classifiers with the merged classifier generated with HM3 and DARE-TIES (model search).}
\label{fig:bert-large-uncased-phishing-sentiment__dare-ties__accuracy}
\end{figure}

\subsection{Confusion Matrices}\label{sub:confusion_matrices}

\begin{figure}[H]
\centering
\includegraphics[width=1.0\linewidth]{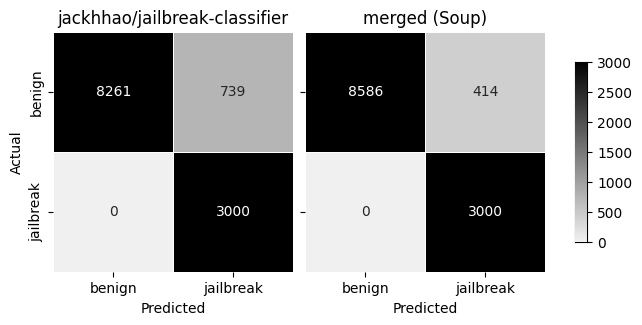}
\caption{Confusion matrix for the jailbreak classifier compared to \methodname-soup.}
\label{fig:bert-base-uncased__linear__jailbreak-detection__confusion_matrices__simple}
\end{figure}

\begin{figure}[H]
\centering
\includegraphics[width=1.0\linewidth]{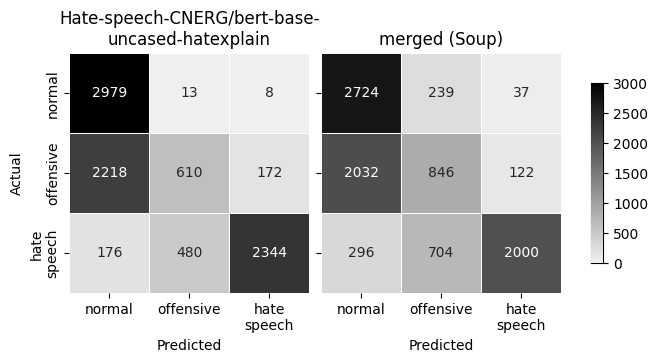}
\caption{Confusion matrix for the hate speech classifier compared to \methodname-soup.}
\label{fig:bert-base-uncased__linear__hate-speech-detection__confusion_matrices__simple}
\end{figure}

\begin{figure}[H]
\centering
\includegraphics[width=1.0\linewidth]{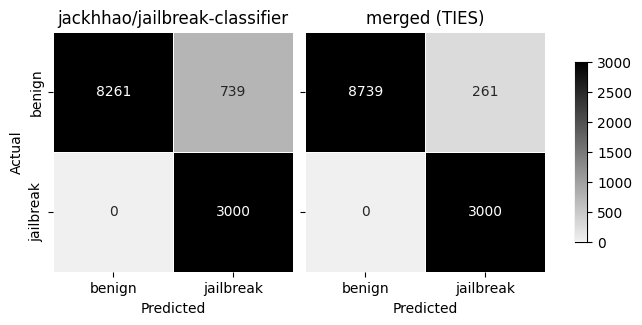}
\caption{Confusion matrix for the jailbreak classifier compared to \methodname-TIES.}
\label{fig:bert-base-uncased__ties__jailbreak-detection__confusion_matrices__simple}
\end{figure}

\begin{figure}[H]
\centering
\includegraphics[width=1.0\linewidth]{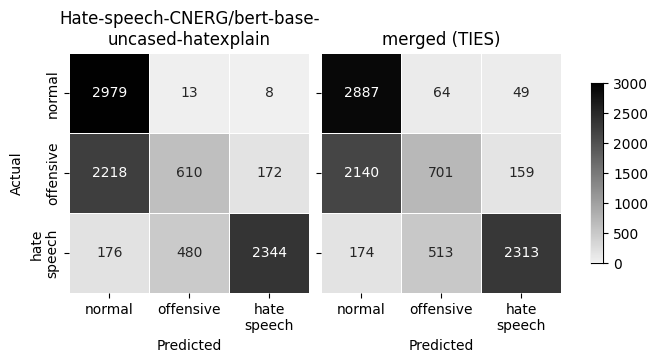}
\caption{Confusion matrix for the hate speech classifier compared to \methodname-TIES.}
\label{fig:bert-base-uncased__ties__hate-speech-detection__confusion_matrices__simple}
\end{figure}

\begin{figure}[H]
\centering
\includegraphics[width=1.0\linewidth]{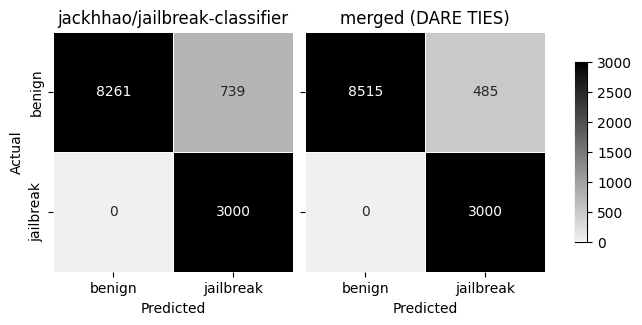}
\caption{Confusion matrix for the jailbreak classifier compared to model search using \methodname-DARE-TIES.}
\label{fig:bert-base-uncased__dare-ties__jailbreak-detection__confusion_matrices__simple}
\end{figure}

\begin{figure}[H]
\centering
\includegraphics[width=1.0\linewidth]{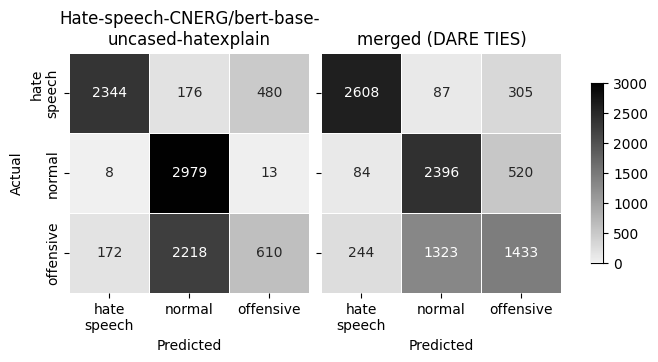}
\caption{Confusion matrix for the hate speech classifier compared to model search using \methodname-DARE-TIES.}
\label{fig:bert-base-uncased__dare-ties__hate-speech-detection__confusion_matrices__simple}
\end{figure}

 \begin{figure}[H]
 \centering
 \includegraphics[width=1.0\linewidth]{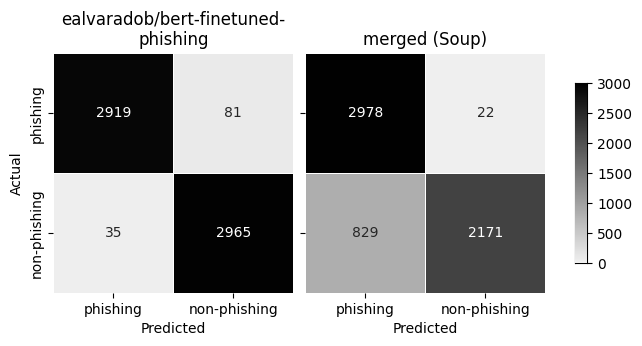}
 \caption{Confusion matrix for the phishing classifier compared to \methodname-soup.}
 \label{fig:bert-large-uncased__soup__phishing-detection__confusion_matrices__simple}
\end{figure}

\begin{figure}[H]
\centering
\includegraphics[width=1.0\linewidth]{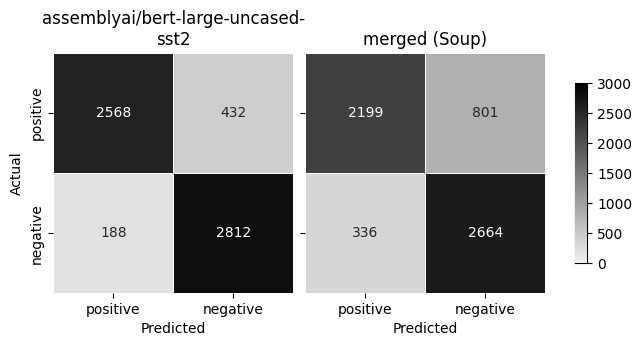}
\caption{Confusion matrix for the sentiment classifier compared to \methodname-soup.}
\label{fig:bert-large-uncased__soup__sentiment-detection__confusion_matrices__simple}
\end{figure}

\begin{figure}[H]
\centering
\includegraphics[width=1.0\linewidth]{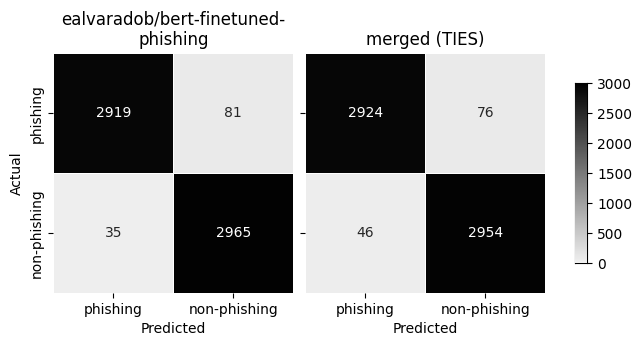}
\caption{Confusion matrix for the phishing classifier compared to \methodname-TIES.}
\label{fig:bert-large-uncased__ties__phishing-detection__confusion_matrices__simple}
\end{figure}

\begin{figure}[H]
\centering
\includegraphics[width=1.0\linewidth]{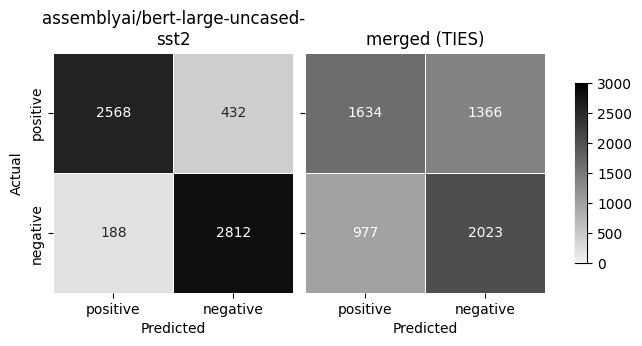}
\caption{Confusion matrix for the sentiment classifier compared to \methodname-TIES.}
\label{fig:bert-large-uncased__ties__sentiment-detection__confusion_matrices__simple}
\end{figure}

\begin{figure}[H]
\centering
\includegraphics[width=1.0\linewidth]{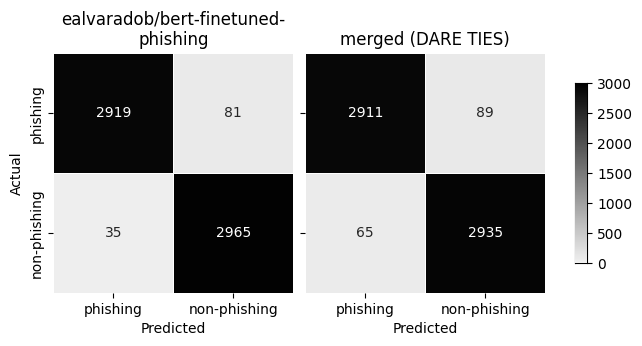}
\caption{Confusion matrix for the phishing classifier compared to model search using \methodname-DARE-TIES.}
\label{fig:bert-large-uncased__dare-ties__phishing-detection__confusion_matrices__simple}
\end{figure}

\begin{figure}[H]
\centering
\includegraphics[width=1.0\linewidth]{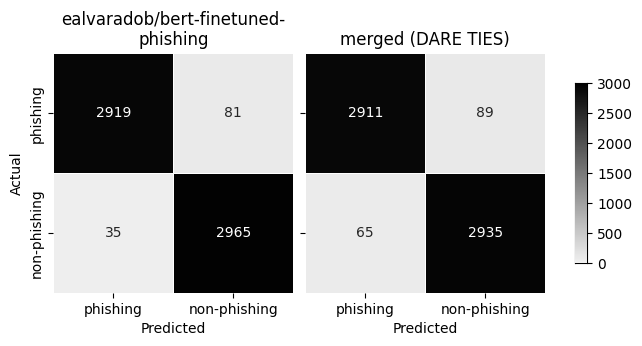}
\caption{Confusion matrix for the sentiment classifier compared to model search using \methodname-DARE-TIES.}
\label{fig:bert-large-uncased-phishing-sentiment__dare-ties__sentiment-detection__confusion_matrices__simple}
\end{figure}

\subsection{Self-Merging}\label{sub:selfmerging_results}
\begin{figure}[H]
    \centering
    \includegraphics[width=1.0\linewidth]{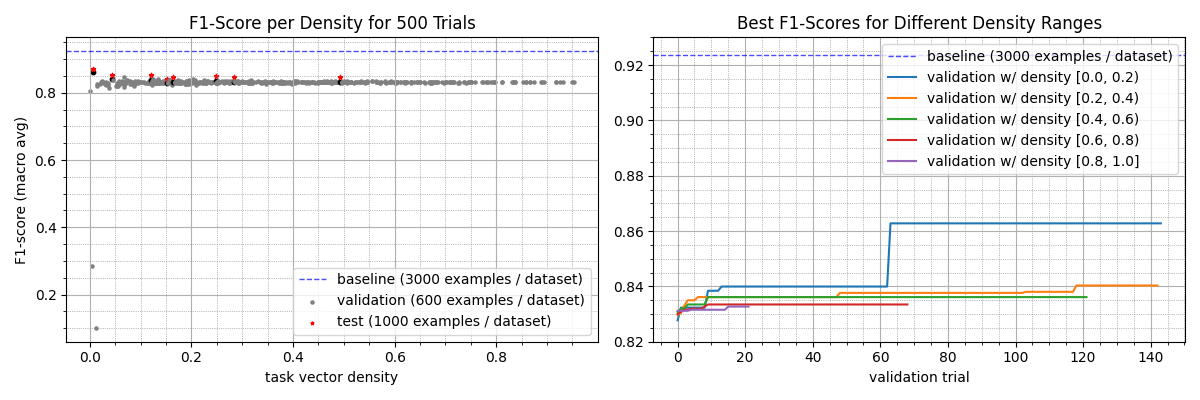}
    \caption{Merging \texttt{jackhhao/jailbreak-classifier} with itself using DARE-TIES, 500 times.}
    \label{fig:selfmerge_jailbreak}
\end{figure}
\begin{figure}[H]
    \centering
    \includegraphics[width=1.0\linewidth]{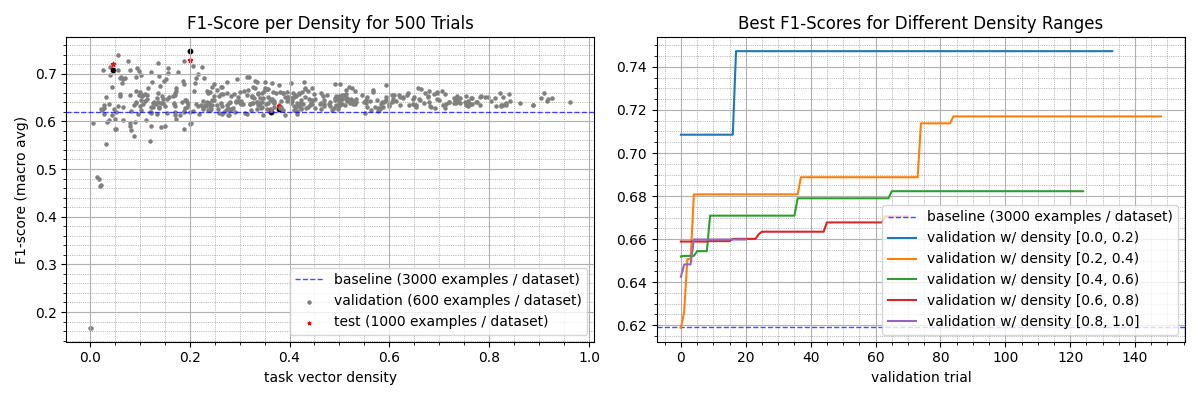}
    \caption{Merge of \texttt{Hate-speech-CNERG/bert-base-uncased-hatexplain} with itself using DARE-TIES, 500 times.}
    \label{fig:selfmerge_hatespeech}
\end{figure}
\begin{figure}[H]
    \centering
    \includegraphics[width=1.0\linewidth]{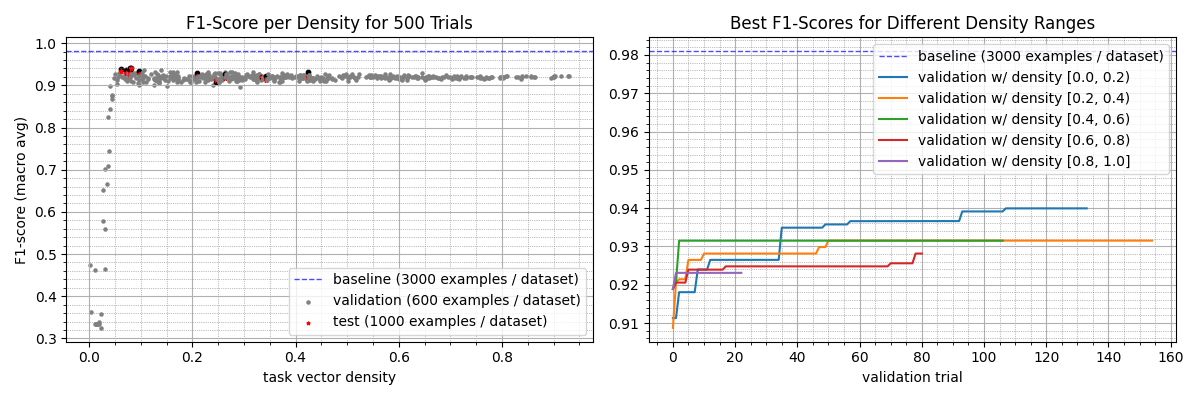}
    \caption{Merge of \texttt{ealvaradob/bert-finetuned-phishing} with itself using DARE-TIES, 500 times.}
    \label{fig:selfmerge_phishing}
\end{figure}
\begin{figure}[H]
    \centering
    \includegraphics[width=1.0\linewidth]{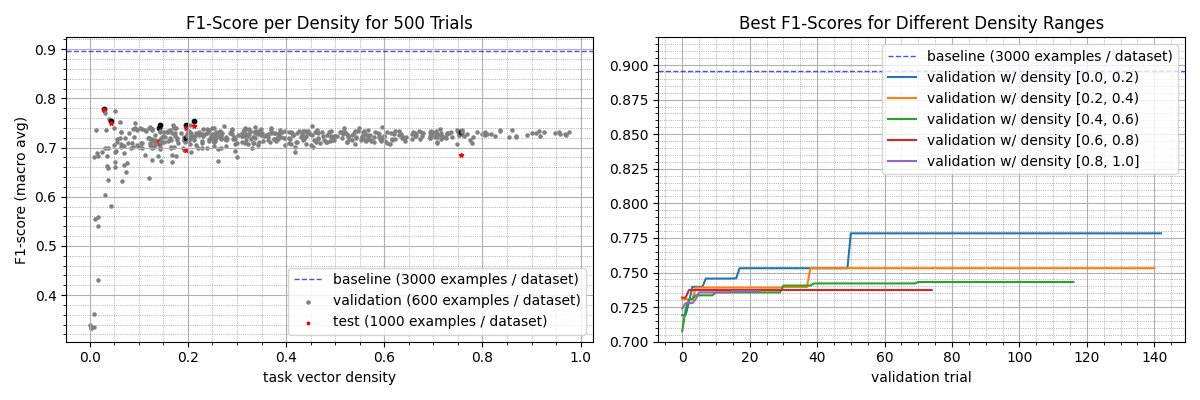}
    \caption{Merge of \texttt{assemblyai/bert-large-uncased-sst2} with itself using DARE-TIES, 500 times.}
    \label{fig:selfmerge_sentiment}
\end{figure}

\subsection{Testing a Merged Text-Classifier}\label{sub:testing}
\begin{figure}[H]
    \centering
    \includegraphics[width=1.0\linewidth]{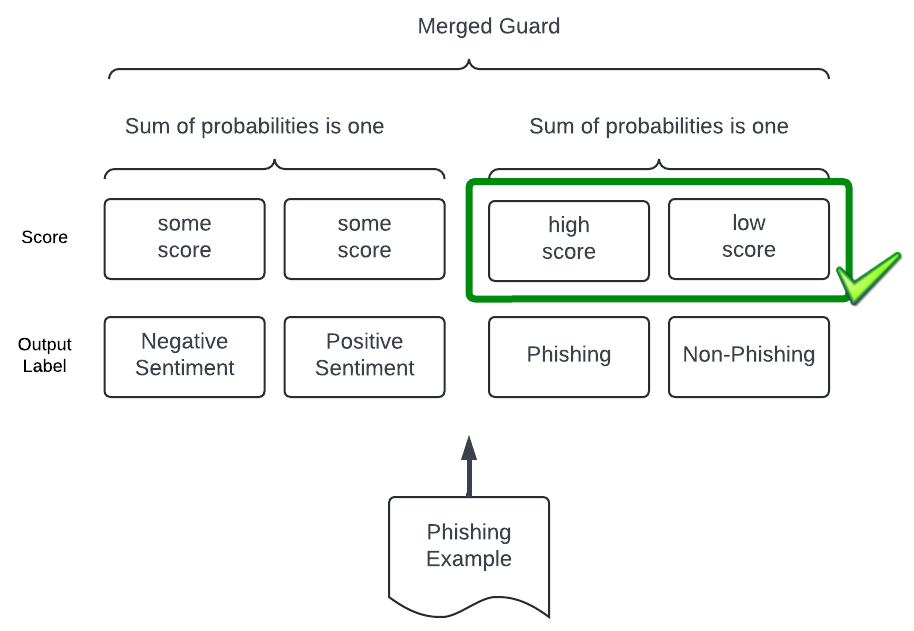}
    \caption{When testing a merged text-classifier, for instance, this merged guard model, we would like to evaluate our test datasets an all available output labels. Where this is possible, we compare the performance of the merged model with the performance of the corresponding original model. However, in some situations, we do not know the expected label and the corresponding test field remains blank. For instance, we could find that the output label for phishing shows a high score/probability when giving a phishing example and use this to compare with the corresponding original model but we do not know how to interpret the sentiment scores in this case.}
    \label{fig:more_explaining}
\end{figure}

\end{document}